\documentclass{ecai} 
\usepackage{latexsym}
\usepackage{amssymb}
\usepackage{amsmath}
\usepackage{amsthm}
\usepackage{booktabs}
\usepackage{enumitem}
\usepackage{graphicx}
\usepackage{color}
\usepackage{algpseudocode}
\usepackage{algorithm}
\usepackage{caption}
\usepackage{subcaption}
\usepackage{hyperref}

\usepackage[bottom]{footmisc}

\newcommand{\BibTeX}{B\kern-.05em{\sc i\kern-.025em b}\kern-.08em\TeX}

\begin{document}

%%%%%%%%%%%%%%%%%%%%%%%%%%%%%%%%%%%%%%%%%%%%%%%%%%%%%%%%%%%%%%%%%%%%%%%%

\begin{frontmatter}

%%% Use this command to specify your submission number.
%%% In doubleblind mode, it will be printed on the first page.

\paperid{123} 

%%% Use this command to specify the title of your paper.

\title{Online hierarchical partitioning of the output space \\ in extreme multi-label data streams}

%%% Use this combinations of commands to specify all authors of your 
%%% paper. Use \fnms{} and \snm{} to indicate everyone's first names 
%%% and surname. This will help the publisher with indexing the 
%%% proceedings. Please use a reasonable approximation in case your 
%%% name does not neatly split into "first names" and "surname".
%%% Specifying your ORCID digital identifier is optional. 
%%% Use the \thanks{} command to indicate one or more corresponding 
%%% authors and their email address(es). If so desired, you can specify
%%% author contributions using the \footnote{} command.

\author[A]{\fnms{Lara}~\snm{Neves}\orcid{0009-0000-3530-1848}\thanks{Corresponding Author. Email: lspsn@isep.ipp.pt.}\footnote{Equal contribution.}}

\author[A]{\fnms{Afonso}~\snm{Lourenço}\orcid{0000-0002-3465-3419}\thanks{Corresponding Author. Email: fonso@isep.ipp.pt	.}\footnotemark}

\author[B]{\fnms{Alberto}~\snm{Cano}\orcid{0000-0001-9027-298X}} 
\author[A]{\fnms{Goreti}~\snm{Marreiros}\orcid{0000-0003-4417-8401}} 
\address[A]{GECAD, ISEP, Polytechnic of Porto, Rua Dr. António Bernardino de Almeida, Porto, 4249-015, Portugal}
\address[B]{Virginia Tech, Virginia Polytechnic Institute and State University, Blacksburg, VA 24061 USA}
% \address{}

\begin{abstract}
Mining data streams with multi-label outputs poses significant challenges due to evolving distributions, high-dimensional label spaces, sparse label occurrences, and complex label dependencies. Moreover, concept drift affects not only input distributions but also label correlations and imbalance ratios over time, complicating model adaptation. To address these challenges, structured learners are categorized into local and global methods. Local methods break down the task into simpler components, while global methods adapt the algorithm to the full output space, potentially yielding better predictions by exploiting label correlations. This work introduces iHOMER (Incremental Hierarchy Of Multi-label Classifiers), an online multi-label learning framework that incrementally partitions the label space into disjoint, correlated clusters without relying on predefined hierarchies. iHOMER leverages online divisive-agglomerative clustering based on \textit{Jaccard} similarity and a global tree-based learner driven by a multivariate \textit{Bernoulli} process to guide instance partitioning. To address non-stationarity, it integrates drift detection mechanisms at both global and local levels, enabling dynamic restructuring of label partitions and subtrees. Experiments across 23 real-world datasets show iHOMER outperforms 5 state-of-the-art global baselines, such as MLHAT, MLHT of Pruned Sets and iSOUPT, by 23\%, and 12 local baselines, such as binary relevance transformations of kNN, EFDT, ARF, and ADWIN bagging/boosting ensembles, by 32\%, establishing its robustness for online multi-label classification.
%Source code, datasets, and results are made publicly available.
\end{abstract}

\end{frontmatter}

%%%%%%%%%%%%%%%%%%%%%%%%%%%%%%%%%%%%%%%%%%%%%%%%%%%%%%%%%%%%%%%%%%%%%%%%

\section{Introduction}

Data stream mining aims to learn models from evolving data, where concept drift alters the underlying distribution over time \cite{gama2010knowledge}. Traditionally, each instance is assumed to have a unique label, but real-world objects often carry multiple, interdependent meanings. This gives rise to streaming multi-label learning (SMLL), where the goal is to predict sets of labels while accounting for label dependencies \cite{read2012scalable}. SMLL faces three major challenges: (1) exponential output space; (2) class imbalance, both within and across labels; and (3) concept drift, which dynamically changes label definitions and correlations \cite{aguiar2024survey}. These challenges are especially acute in streaming settings, where models must adapt continuously and efficiently.

\begin{figure}[H]
    \centering
    \includegraphics[width=0.5\textwidth]{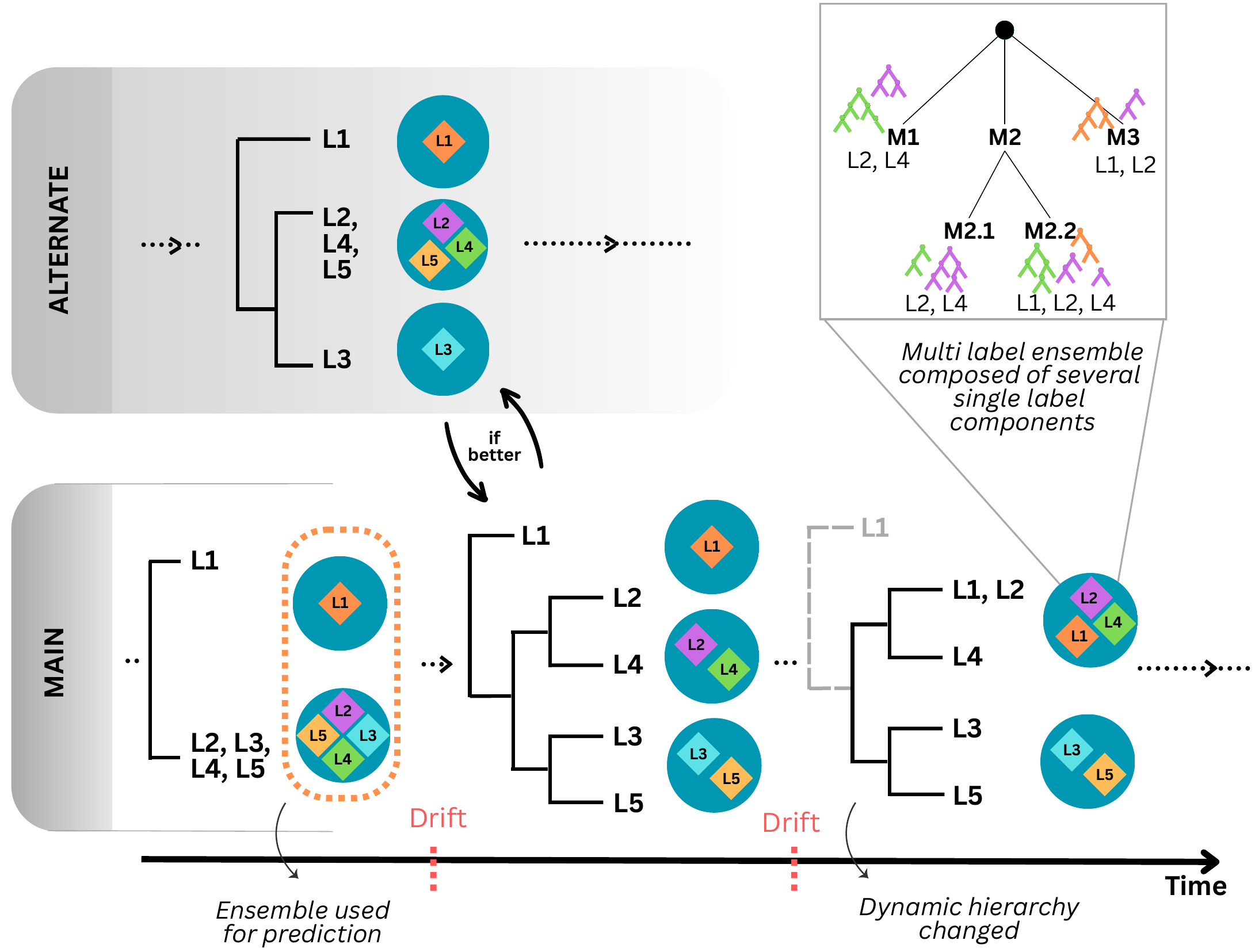}
    \caption{Hierarchy of label clustered classifiers evolves over time.}
    \label{fig:initial}
\end{figure}

\textbf{Local vs. Global Models.} Structured prediction methods are typically divided into local and global approaches \cite{kocev2007ensembles}. Local methods decompose multi-label tasks into independent scalar predictions, while global methods train a single model to output structured predictions directly \cite{duarte2015multi, osojnik2017multi}, often yielding superior performance by better capturing label dependencies. To combine the strengths of both, output specialization methods model label subsets with global predictors, using label space partitioning to create clusters that lie between fully local and global extremes \cite{sousa2018multi}. HOMER \cite{tsoumakas2008effective} uses hierarchical decomposition to simplify learning in high-dimensional spaces, while more recent work like HPML \cite{gatto2025multi} explores flat, non-hierarchical clusters to balance accuracy and efficiency.

\textbf{This work.} Building on these ideas, this work proposes iHOMER (Incremental Hierarchy Of Multi-label classifiERs), an online approach that clusters the label space into disjoint, correlated groups, forming hybrid partitions that combine global and local strategies (Figure \ref{fig:initial}). iHOMER does not rely on fixed hierarchies and avoids expensive tree construction while preserving partition coherence. Experiments on 23 real-world datasets show iHOMER outperforms 5 global baselines by 23\%, 12 local baselines by 32\%, and random search by 40\%, validating its effectiveness. Source code, datasets, and results are publicly available \footnotemark[2].

%Unlike HOMER, iHOMER does not rely on fixed hierarchies, and like HPML, it avoids expensive tree construction while preserving partition coherence. Experiments on 23 real-world datasets show iHOMER outperforms 5 global baselines by 23\%, 12 local baselines by 32\%, and random search by 40\%, validating its effectiveness. Source code, datasets, and results are publicly available.

%%%%%%%%%%%%%%%%%%%%%%%%%%%%%%%%%%%%%%%%%%%%%%%%%%%%%%%%%%%%%%%%%%%%%%%%

\section{Related work}

Multi-label learning algorithms model label correlations in various ways, which can be categorized along four dimensions: (1) instance location (global vs. local), (2) dependency (conditional vs. unconditional), (3) dataset space (feature, instance, or label), and (4) correlation order (first-, second-, or high-order) \cite{zhang2007ml}. Global correlations consider the full dataset, while local ones work within subsets like clusters. Conditional correlations depend on instance features, whereas unconditional ones model co-occurrence across all data. Correlations may arise from the feature, instance, or label space. First-order models ignore label dependencies, e.g., Binary Relevance (BR), second-order models capture pairwise relations, and high-order methods model interactions among multiple labels. Despite these differences, these multi-label strategies share patterns: global/high-order/unconditional methods model the full label set, while local/second-order/conditional methods target subsets. More interestingly, these can be grouped into three algorithmic families: local models, global models, and hybrid partitions.

\textbf{Local approaches.} Problem transformation methods fit the data for traditional single-label streaming classifiers. In particular, BR is commonly used for its scalability and parallelism \cite{boutell2004learning}, though it suffers from class imbalance and empty predictions. Streaming BR solutions include label-wise weighting \cite{raez2004adaptive}, separate windows for positives and negatives \cite{spyromitros2011dealing}, and concept drift detection via label entropy \cite{tomas2014framework}. MINAS-BR handles novelty and unlabeled labels in extreme streaming settings \cite{junior2023novelty}. Beyond BR, second-order methods like Label Ranking use one-vs-one classifiers to model pairwise label orderings, though scalability is an issue in high dimensions \cite{madjarov2012two}. High-order methods like Label Combination treat label sets as multi-class targets \cite{shi2014drift}, while managing the growing number of label sets with probabilistic modeling \cite{wei2021probabilistic} or early-sample buffering \cite{read2012scalable}.

\textbf{Global approaches.} Algorithm adaptation methods directly handle multi-label outputs, reducing complexity and making them ideal for streaming. Notable examples include extensions of ML-kNN \cite{zhang2007ml}, using Bayesian inference over neighbors, and extensions of ML-DT \cite{clare2001knowledge}, using multi-label entropy. Streaming ML-kNN solutions include windowing to efficiently update posteriors \cite{spyromitros2011dealing} and dual-memory models like ML-SAM-kNN \cite{roseberry2018multi}, which splits recent and long-term data. Extensions such as MLSAMPkNN \cite{roseberry2019multi} and MLSAkNN \cite{roseberry2021self} introduce punitive mechanisms and adaptive clustering. ARkNN \cite{roseberry2023aging} proposes resource-efficient instance aging. Streaming ML-DT solutions extend Hoeffding Trees (HT) \cite{domingos2000mining} with label-set predictors and pruning techniques. For example, iSOUPT \cite{osojnik2017multi} applies multi-target regression via perceptron-based model trees, which can be extended with drift detection \cite{stevanoski2024change} and per-target model selection \cite{mastelini2019online}. MLHAT \cite{esteban2024hoeffding} goes beyond, incorporating drift and label imbalance adaptation via cardinality-aware metrics.

\textbf{Hybrid partitions.} It is important to note that the aforementioned global approaches are sometimes referred to as ensemble methods due to their use of multiple binary models. Similarly, hybrid partitioning methods are occasionally labeled as ensemble methods because they involve several global multi-label models. However, since these models handle disjoint label subsets and collectively represent a unified prediction system, they effectively constitute a single multi-label model. Hybrid partition methods blend local and global models by clustering labels and applying global predictors to each subset \cite{sousa2018multi}, offering a middle ground between fine-grained and holistic learning. By modeling label correlations locally, i.e., within label clusters \cite{huang2015group}, or across all labels \cite{gatto2025multi}, these methods aim to efficiently guide label space partitions, rather than relying on random partitioning \cite{szymanski2016data,breskvar2018ensembles} or genetic algorithms that overlook label dependencies \cite{basgalupp2021beyond,moyano2020combining}. In fact, examples of this can be found beyond tree-based methods, leveraging the modular nature of rule learning systems to build localized models over partitions of the input space. For example, AMRules \cite{almeida2013adaptive} has been extended to multi-label classification \cite{duarte2016adaptive}, incorporating mechanisms for handling concept drift using the Page-Hinkley test and detecting anomalous instances. This modularity enables surpassing both global and local correlation-based methods via rule specialization for non-disjoint outputs in multi-target regression \cite{duarte2015multi} and multi-label classification \cite{sousa2018multi}. When performing rule expansion, if the variance in the target space is reduced only for certain targets, a new rule is created for those targets, while a complementary rule is retained for the rest. As a result, the system can generate rules covering all, some, or just a single target.

\textbf{iHOMER.} Building on these advancements, iHOMER is the first incremental tree-based output specialization approach, whose main novelty comes from 4 distinct characteristics: (1) flexibility, (2) non-repetition, (3) computational efficiency, and (4) adaptive mechanisms. Firstly, while some algorithms require high parameterization for searching the best output space partition, such as defining the number of individuals, the number of generations, the number of classifiers, and the number of labels for each classifier in genetic algorithms \cite{basgalupp2021beyond,moyano2020combining}, iHOMER allows the creation of various hybrid partition configurations without the need to previously determine their number. Secondly, iHOMER ensures that labels are not repeated in label clusters for each hybrid partition generated by adjusting cluster labels to satisfy this constraint, which is not done in related methods \cite{papanikolaou2018hierarchical,szymanski2016data}. Thirdly, iHOMER leverages extremely resource-efficient algorithm choices, e.g., a growing clustering structure based on an incremental \textit{Jaccard} dissimilarity measure, and a growing tree structure based on a multivariate \textit{Bernoulli} process, both of which are statistically guided by the \textit{Hoeffding bound} \cite{domingos2000mining}. Fourthly, iHOMER incorporates drift detection mechanisms at both global and local levels, enabling dynamic restructuring of label partitions and subtrees, validated with an alternate background replacement process \cite{bifet2009adaptive}.

\footnotetext[2]{\url{http://github.com/larasneves/iHomer}}

\section{Methodology}

iHOMER is an online approach that clusters the label space to create hybrid partitions. Firstly, correlations between labels are modeled at a dataset level with the \textit{Jaccard} similarity measure. Secondly these are clustered in order to obtain hybrid partitions of the label space, with a top-down growing hierarchy of splits governed either by a user-defined threshold on the number of observations, denoted \( n_{\min} \), or by a statistically grounded criterion based on the \textit{Hoeffding bound}. Thirdly, as the data distribution evolves, rendering earlier partitioning decisions obsolete, previous splits and child nodes are re-aggregated into their parent node, also based on the \textit{Hoeffding bound}. Moreover, since iHOMER should extract the most meaningful and balanced clusters from the hierarchy at any given time, a second-stage balanced aggregation is performed when necessary, according to a minimum safe cluster size determined via a majority Labelset classifier analyzing the distribution of labelset sizes. Subsequently, a drift detection mechanism monitors the multi-label accuracy, allowing it to trigger an alternate clustering process in the background, which replaces the main one if the concept drift is confirmed. Finally, a global multi-label incremental tree-based learner is applied for each constructed partition. It is important to note that since this ensemble of global models is concerned with disjoint sets of labels, iHOMER in fact constitutes a single multi-label model.

\subsection{Incremental dissimilarity measure}

\begin{figure}
    \centering
    \includegraphics[width=0.40\linewidth]{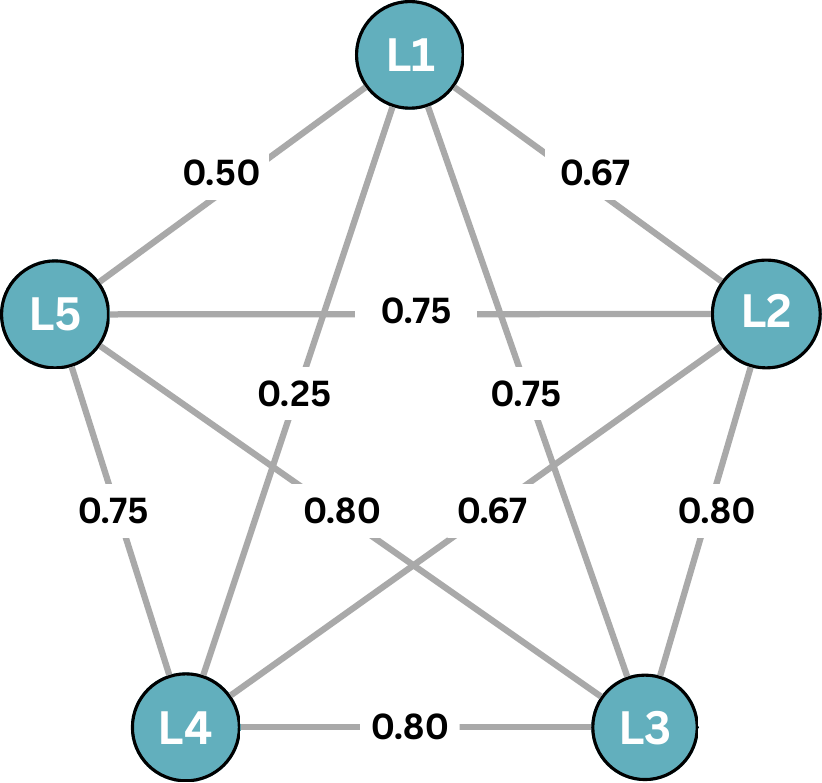}
    \caption{Dissimilarity graph based on label co-occurrences.}
    \label{dissimilarity}
\end{figure}

To effectively capture pairwise similarity degrees of all labels in the label space, the \textit{Jaccard} index was used \cite{gatto2025multi}. Formally, considering \( p_j \) and \( q_j \) as two distinct labels within the label space, the label dissimilarity is defined as:

\begin{equation}
d(\text{Jaccard}) = (1 - \frac{a}{a + b + c})
\end{equation}

where \( a \) represents the number of instances simultaneously classified with both \( p_j \) and \( q_j \); \( b \) denotes the number of instances classified only with \( p_j \); and \( c \) is the number of instances classified only with \( q_j \). It is important to note that the factors used to compute the \textit{Jaccard} index can be updated incrementally. As each example is processed once, sufficient statistics are updated incrementally, while the dissimilarity graph, shown in Figure \ref{dissimilarity}, is only computed when testing for splitting or aggregation. Missing labels are imputed as zero.

\subsection{Growing output space hierarchy}

\begin{figure}
    \centering
    \includegraphics[width=0.85\linewidth]{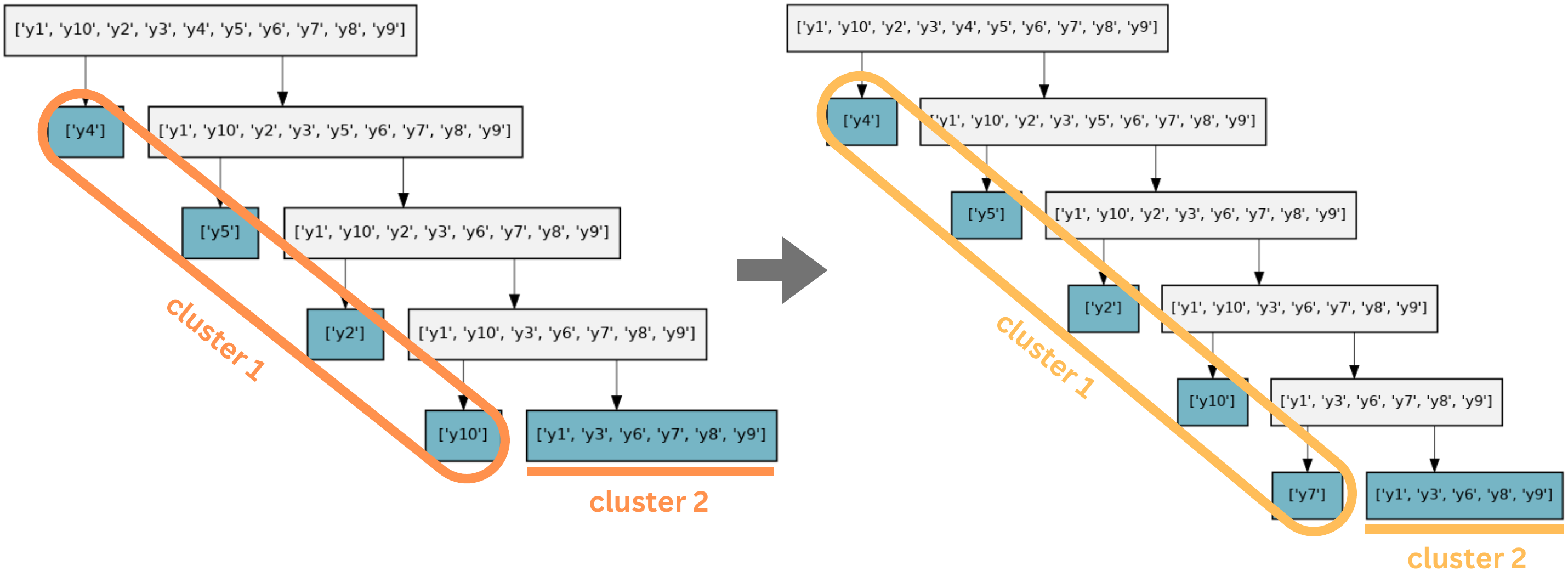}
    \caption{Dynamic hierarchical label clusters in Hypersphere dataset.}
    \label{merging}
\end{figure}

The iHOMER algorithm employs online divisive agglomerative clustering to incrementally construct a tree-structured hierarchy that reflects the nested organization of label clusters. This process follows a top-down strategy, as illustrated in Figure \ref{merging}. The leaves of the tree correspond to the resulting clusters, each associated with a distinct subset of variables~\cite{odac}. The union of all leaves comprises the complete variable set, and the intersection between any two leaves is empty. During the growing process, based on the incremental dissimilarity measure and the corresponding diameters of the clusters, a key decision is made on whether a given leaf node, i.e., cluster, should be split. This cluster splitting is governed either by a user-defined threshold on the number of observations, denoted \( n_{\min} \), or by a statistically grounded \textit{Hoeffding bound} criterion, which provides a concentration inequality for the mean of a bounded random variable. Specifically, for a real-valued random variable \( r \) bounded in range \( R \), after observing \( n \) independent samples, the bound ensures that, with confidence \( 1 - \delta_{cluster} \), the true mean lies within \( \epsilon \) of the empirical mean, where:

\begin{equation}
\epsilon = \sqrt{\frac{R^2 \ln(1/\delta_{cluster})}{2n}}.
\end{equation}

Each cluster \( C_k \) maintains its own Hoeffding threshold \( \epsilon_k \) to determine whether the diameter-based heuristic condition is met. In particular, the splitting criterion compares the spread of pairwise dissimilarities within the cluster to the deviation from the mean:

\begin{equation}
\frac{(d_1 - d_0)}{|d_1 + d_0 - 2d|} > \epsilon_k.
\end{equation}

where \( d_1 \) is the maximum distance between any two elements in the cluster, \( d_0 \) is the minimum pairwise distance, and \( d \) is the average dissimilarity across all pairs. If this inequality holds, the cluster is considered sufficiently diverse to justify splitting. In this case, the observations corresponding to the pair with maximum dissimilarity are selected to initiate new clusters as pivot variables, and the remaining variables are assigned to the nearest new cluster based on pivot proximity. This results in two subclusters centered around the most dissimilar elements in the original cluster. While this process enables the iHOMER tree to grow in a data-driven and statistically sound manner, maintaining robustness against noise and avoiding over-segmentation, near-equidistant variables can delay the decision between two equally valuable splits. To address this, a parameter $\tau_{cluster}$ is introduced. If $\tau_{cluster} > \epsilon_k$, the system forces a test, assuming enough examples have been seen.

\subsection{Aggregating output space hierarchy}

A critical feature of iHOMER lies in its continuous monitoring of cluster diameters, which provides a basis not only for initiating splits but also for detecting when prior splits may no longer reflect the current structure of the data. In stationary data streams, each successive split within a divisive hierarchical clustering scheme typically leads to a reduction in intra-cluster dissimilarity, producing increasingly refined subgroups. However, in non-stationary environments, the data distribution may evolve, rendering earlier partitioning decisions obsolete. In such cases, the system must possess the flexibility to reverse previous splits and re-aggregate child nodes into their parent. To detect when a previous split has become inadequate, the algorithm assesses whether the diameters of sibling clusters remain meaningfully smaller than that of their parent:

\begin{equation}
2 \cdot \text{diam}(C_j) - (\text{diam}(C_k) + \text{diam}(C_s)) < \epsilon_j.
\end{equation}

where \( C_k \) is a given leaf node, \( C_s \) its sibling, \( C_j \) their common parent, and \( \epsilon_j \) the Hoeffding-based confidence interval derived from the amount of data previously accumulated at the parent node. If this condition holds, then there is insufficient evidence to justify maintaining the split, and the system chooses to re-aggregate \( C_k \) and \( C_s \) into their parent \( C_j \). The resulting re-aggregated node begins with fresh statistical summaries, under the assumption that only future observations will inform its internal structure. This initialization allows the system to remain responsive to evolving data patterns. Moreover, the dual use of the \textit{Hoeffding bound}, for both splitting and re-aggregation, enhances iHOMER’s ability to adapt to changing data distributions while maintaining a balance between sensitivity to novel patterns and stability against noise. However, the resulting clusters might be unbalanced. Ideally, iHOMER should extract the most meaningful and balanced clusters from the hierarchy at any given time, as illustrated in Figure \ref{merging}. To maintain flexibility of the hierarchy building process, a second-stage aggregation is performed only when necessary, ensuring clusters represent coherent groups rather than isolated individual labels. Leveraging on the fact that the produced dendrogram can be cut at different levels, this process adjusts the tree structure by merging clusters when appropriate, refining the representation of the data over time. To achieve this, the Majority Labelset classifier is used, when requested, to analyze the distribution of labelset sizes. It retrieves a frequency table indicating how often labelsets with a given number of labels occur, using this information to determine the minimum safe cluster size. This approach is intuitive as it suggests that clusters should contain at least as many labels as the most common labelset size. For example, if the most frequent labelset size in a dataset is eight, predicting only two labels at a time would be inefficient. Conversely, if most instances have just one label, maintaining specialized models for single-label prediction is more effective. Since this strategy follows a hierarchical clustering approach, imposing a maximum number of labels per cluster is unnecessary and may hinder the natural refinement of the structure over time. Theoretically, all labels could initially form a single cluster, but as new observations are processed, the hierarchy self-adjusts, splitting or merging clusters as needed. This adaptive process ensures that the clustering structure evolves to balance granularity and coherence, making rigid constraints on cluster size unnecessary. 

\subsection{Alternate output space partitioning}

As the data distribution evolves, rendering earlier partitioning decisions obsolete, iHOMER dynamically aggregates or expands its label partitioning, allowing faster recovery rates. However, the base learner performance still requires time to improve with the continuous arrival of new samples, as the historical data before the concept drift plays a restraining role in the model’s update, leading to the inability of classification accuracy to recover in a short time. In other words, if the learning model has been in a state of fail-safe operation, i.e., carrying the original distributed samples or information, for a long time, it might not be able to adapt to the demand of high timeliness in streaming data. Furthermore, in addition to true concept drifts, iHOMER is sensitive to false concept drifts, such as deviation of noise samples from the normal distribution, and unstable representations due to updated small-scale datasets. Consequently, to smooth these transitions, iHOMER features a conservative replacement strategy, by selectively retiring outdated models \cite{bifet2009adaptive}. When iHOMER detects multiple consecutive drift signals, suggesting the active learner may be misaligned with the data stream, an alternative learner is initialized. This alternate component explores a new representation of the data stream and can replace the active model only if it demonstrates superior performance in terms of current subset accuracy. Specifically, the \textit{Welch’s T-test} is applied to assess whether the prediction errors $\alpha, \alpha_{c'}$ of the active label clustering strategy \( c \) and a candidate alternative \( c' \) differ significantly, under the assumption of unequal variances. The test statistic \( t_{statistic} \), of confidence $1-\delta_{alt-cluster}$, is computed as:

\begin{equation}
t_{statistic} = \frac{\alpha_c - \alpha_{c'}}{\sqrt{\frac{\sigma_c^2}{N_c} + \frac{\sigma_{c'}^2}{N_{c'}}}}
\label{t_stat}
\end{equation}

where \( \epsilon \) denotes the mean error, \( \sigma^2 \) the error variance, and \( N \) the number of instances observed in an \textit{ADWIN}-controlled sliding window for each model \cite{bifet2009adaptive}. When the test indicates a statistically significant difference, the alternative partitioning may replace its counterpart, supporting dynamic adaptation to concept drift in the data stream. Moreover, recursive replacements are possible, with multiple competing alternatives, specializing in different temporal regions of the data stream. The more recent alternate learner typically adapts quickly to current trends, making it particularly effective when dealing with abrupt concept drifts. In contrast, the older and generally more stable active learner adapts more slowly but tends to perform better in stationary or gradually evolving segments of the stream. This mechanism enables dynamic retirement of underperforming active models while preserving stability. Moreover, the introduction of a \textit{grace period} $n_{\min}$ before replacement decisions helps reduce false alarms and stabilizes system responsiveness in the face of noise and short-term fluctuations. 

\subsection{Global base learner}

Once iHOMER has identified meaningful label clusters through hierarchical partitioning, a specialized multi-label Hoeffding adaptive tree (MLHAT) is trained on each cluster. Following the principles of incremental Hoeffding trees \cite{domingos2000mining}, MLHAT leverages on each incoming instance, updating sufficient statistics from the root to a leaf. For every attribute-value-class combination \( n_{ijk} \), the algorithm tracks how often an attribute \( X_i = x_{ij} \) appears with class \( y_k \), enabling evaluation of potential splits. The base learner then chooses the best attribute using a heuristic function \( G(\cdot) \), i.e., the information gain based on the \textit{Bernoulli} distribution, where label priors are obtained from the counters maintained by the node \cite{esteban2024hoeffding}. It compares the top two attributes and applies the \emph{Hoeffding Bound} to decide whether the difference \( \Delta G \) is statistically significant, with confidence $1-\delta_{tree}$:

\begin{equation}
    \epsilon_{m} = \sqrt{\frac{\log_2(|L|)^2 \ln(1/\delta_{tree})}{2n}}
\end{equation}

where \( L \) is the number of known label sets, \( \delta \) is the desired confidence level, and \( n \) is the number of samples at the node $m$. A split occurs when \( \Delta G > \epsilon_{h} ,\). In this way, the tree would expand incrementally as long as the entropy of the leaf nodes can be minimized with new splits. To manage complexity and prevent overfitting in incremental decision trees, two key hyperparameters are used: the \textit{grace period} \(n_{\min}\), which defines how often splits are evaluated, and the \textit{tie threshold} \(\tau_{tree}\), which serves as a pre-pruning mechanism. The tie threshold \(\tau_{tree}\) limits growth when the difference \(\Delta \bar{H}(X_i)\) between the best two attribute heuristics is small, i.e., \(\Delta \bar{H}(X_i) \leq \tau_{tree}\). This helps avoid unstable decisions under noisy or ambiguous data. While a well-chosen \(\tau\) mitigates unnecessary tree growth and improves scalability, its effectiveness is highly data-dependent. If too low, the tree may underfit; if too high, it risks overfitting and computational inefficiency. Together, \(\tau_{tree}\) and \(n_{\min}\) balance accuracy and complexity in data stream learning, allowing the production of increasingly larger trees that can divide the problem space into finer regions, better approximating the target concepts. However, such tree growth still remains highly uncontrollable, which can lead to two main issues: excessive memory consumption due to multiple redundant splits on all features, and a loss of plasticity due to the descendant nodes getting subsequently fixed to the space covered by their parent node. To address this, outdated subtrees are replaced with alternates that are grown when a drift detector detects drift through a changing error rather than on changing split attributes \cite{bifet2009adaptive}. However, such split revision procedure can also reduce variance by simplifying the model, and introduce significant bias if important information is lost during subtree pruning \cite{manapragada2018extremely}. Thus, only when the alternate subtree outperforms the original over a sufficient number of instances does it replace the latter. Following this intuition, MLHAT monitors the error in the Hamming loss, triggering replacements based on the \textit{Hoeffding bound} $\epsilon_{alt}$ for alternating trees \cite{bifet2009adaptive} and defined from monitored errors and instances seen by both main $W$ and alternate $W'$ sub-trees as:

\begin{equation}
    \epsilon_{alt} = \sqrt{\frac{2e (1-e') (W+W') \ln(2/\delta_{alt-tree})}{W \cdot W'}}
    \label{eq:hoeffalt}
\end{equation}

where three scenarios can occur: (1) if $(e - e') > \delta_{alt-tree}$, the alternate tree replaces the main one, pruning the previous sub-structure resetting the drift detector in the alternate tree for future concept drifts, (2) if $(e' - e) > \delta_{alt-tree}$, the alternate tree remains without changes and the alternate tree is pruned, the concept drift was reversed, and (3) if $\delta_{alt-tree} \geq |e - e'|$, the two sub-trees continue to receive instances and expand to eventually differentiate their performance. This process aims to reduce bias and keep the model relevant by ensuring new subtrees adapt to the updated concept before deployment, with recursive replacements among alternates of alternates being possible. In other words, having alternative subtrees switch between training and test modes, and during the latter, also pruning the alternative subtrees whose accuracy does not increase over time. Moreover, leveraging on the fact that, in the midst of a concept drift, there will be multiple alternate sub-trees that are candidates to replace the branches whose performance is declining, their predictions is combined with the main tree, weighted based on the error monitored in each of the leaves involved, and conditioned on the alternate trees having received at least $n_{\min}$ instances. Thus, providing a good balance between the main structure that has the statistical guarantees of the \textit{Hoeffding bound}, and a fast response to possible concept drifts. The full pseudocode for iHOMER is presented in Algorithm \ref{iHOMER}.
\textit{Note:} Parameters are scoped as follows: $\delta$, and $\tau$ for trees use subscript \texttt{tree}; for label clustering, use \texttt{cluster}; and for alternate subtree comparisons, use \texttt{alt-tree}.

\begin{algorithm}[h]
\caption{iHOMER}
\label{iHOMER}
\begin{algorithmic}[1]
\scriptsize

\Statex \textbf{Given:} $n_{\min}$: minimum number of instances per node
\Statex \hspace{0.7cm} $\delta$: confidence levels 
\Statex \hspace{0.7cm} $\tau$: tie-breaking thresholds
\Statex \hspace{0.7cm}  $|L|$: number of known label sets, $G$: heuristic function

\Statex \textbf{Initialize:} cluster hierarchies $H \in \{H_{\text{main}}, H_{\text{alt}}\}$, each composed of models $h$

\Statex 
\For{each $(x, y)$ in stream}
        \For{each cluster hierarchy $H$ of $\{H_{main}, H_{alt}\}$}
            \State Route $x$ to clusters in $H$, and predict disjoint $\hat{y}$
            \State \textit{// Online divisive agglomerative clustering}
            \For{each leaf $l$ in $H$ not yet tested} 
                \State Update d(\text{\textit{Jaccard}})
                \State $\epsilon_l = \sqrt{\frac{R^2 \ln(1/\delta_{cluster})}{2n}}$
                \If{ $\frac{(d_1 - d_0)}{|d_1 + d_0 - 2d|} > \epsilon_l$ and $\epsilon_l < \tau_{cluster}$} 
                %\State{If clustering suggests split/merge}
                    \State Split leaf $l$
                    \State Initiate new clusters as pivot variables
                    \State Assign remaining based on pivot proximity
                \EndIf
                \If{$2 \cdot \text{diam}(C_{parent}) - (\text{diam}(C_l) + \text{diam}(C_{sib})) \geq \epsilon_{\text{l}}$} 
                    \State Re-aggregate \( C_l \) and \( C_{sib} \) into \( C_{parent} \)
                    \State Reinitialize statistics at \( C_{parent} \)
                    \State Generate balanced structure
                \EndIf
            \EndFor
            \For{each model $h$ in cluster hierarchy $H$}
                \For{each node $n$ in the path to $l$} 
                    \State Update node statistics with $(x, y)$
                    \If{alternate subtree $T_{\text{alt}}(n)$ exists}
                        \State Recursively update $T_{\text{alt}}(n)$ with $(x, y)$
                    \EndIf
                    \State Update ADWIN with $(\hat{y}, y)$
                    \If{$T_{\text{alt}}(n)$ exists and $W(T_{\text{alt}}(n)) > n_{\text{min}}$}
                        \State $\epsilon_{alt} = \sqrt{\frac{2e (1-e') (W+W') \ln(2/\delta_{alt-tree})}{W \cdot W'}}$
                        \If{$e - e' > \delta_{alt-tree}$}
                            \State Replace main one with $T_{\text{alt}}(n)$
                        \ElsIf{$e' - e > \delta_{alt-tree}$}
                            \State Prune $T_{\text{alt}}(n)$
                        \EndIf
                    \EndIf
                    \If{ADWIN warns and no $T_{\text{alt}}(n)$ exists}
                        \State $T_{\text{alt}}(n) \gets$ new substitute subtree at $n$
                    \EndIf
                    \If{$n$ is a leaf $l$, and seen instances $> n_{\text{min}}$}
                        \State Evaluate splits using heuristic $G$
                        \State $\epsilon_{l} = \sqrt{\frac{\log_2(|L|)^2 \ln(1/\delta_{tree})}{2n}}$
                        \If{$G_{\text{best}} - G_{\text{sec}} > \epsilon_l$ and $\epsilon_l < \tau_{tree}$}
                            \State Split leaf $l$ on best attribute
                            \State Create child nodes
                        \EndIf
                    \EndIf
                \EndFor
            \EndFor
        \EndFor
    \State Concatenate disjoint $(\hat{y},y)$, and compute errors $\alpha_{main}$, $\alpha_{alt}$
     \If{$\frac{\alpha_{main} - \alpha_{alt}}{\sqrt{\frac{\sigma_{main}^2}{N_{main}} + \frac{\sigma_{alt}^2}{N_{alt}}}} < t_{statistic}$}
        \State Replace the $H_{main}$ with $H_{alt}$ 
    \ElsIf{drifting $H_{main}$}
        \State (Re)initialize $H_{alt}$
    \EndIf
\EndFor
\State \textbf{Output:} Final evaluation metrics
\end{algorithmic}
\end{algorithm}
\section{Experiments}

iHOMER was compared against 19 other multi-label incremental algorithms proposed over the past 25 years, covering a broad spectrum of adaptability to both concept drift and class imbalance. Some of these algorithms address both challenges, while others focus on only one or neither. Within local models, binary relevance transformations of classical incremental decision tree models, like Hoeffding Tree (HT) \cite{domingos2000mining}, Hoeffding Adaptive Tree (HAT) \cite{bifet2009adaptive}, and Extremely Fast Decision Tree (EFDT) \cite{manapragada2018extremely}, were included, as well as more recent approaches such as Stochastic Gradient Tree (SGT) \cite{gouk2019stochastic}, Mondrian Tree (MT) \cite{mourtada2021amf}, Aggregated Mondrian Forest (AMF) \cite{mourtada2021amf}, and Adaptive Random Forest (ARF) \cite{gomes2017adaptive}. Moreover, the comparison spans diverse families of algorithms, including Gaussian Naïve Bayes (NB) \cite{bifet2009new}, k-Nearest Neighbors (kNN) \cite{montiel2021river}, Adaptive Model Rules (AMR) \cite{duarte2016adaptive}, and ensemble methods like ADWIN Bagging of Logistic Regression (ABALR) \cite{bifet2009new} and ADWIN Boosting of Logistic Regression (ABOLR) \cite{bifet2009new}. Within global models, it also includes specialized multi-label methods like Multi-Label Hoeffding Tree (MLHT) \cite{read2012scalable}, MLHT with Prune Set (MLHTPS) \cite{read2012scalable}, Incremental Structured Output Prediction Tree (iSOUPT) \cite{osojnik2017multi}, Multi-Label Broad Ensemble Learning System (MLBELS) \cite{bakhshi2024balancing}, and Multi-Label Hoeffding Adaptive Tree (MLHAT) \cite{esteban2024hoeffding}. To enable a fair and reproducible comparison, the same hyperparameter configurations and evaluation protocols as those reported in the associated open-source repository referenced in the literature \cite{esteban2024hoeffding}. A detailed description of each algorithm’s characteristics, along with the full parameter specifications used in the experiments, is available therein.

\textbf{Evaluation metrics.}  
Given the incremental nature of data streams, the standard prequential evaluation (also known as test-then-train \cite{gama2009evaluating}) was adopted. This approach evaluates the model on each instance before using it for training, allowing for continuous monitoring of predictive quality over time. Following prior work, we selected five widely-used multi-label metrics: \textbf{Subset Accuracy}, \textbf{Sample Accuracy}, \textbf{Hamming Loss}, \textbf{Micro F1}, and \textbf{Macro F1}. This choice ensures a fair comparison with existing methods and captures both instance-level and label-level behavior.To analyze model behavior over time, the rolling Sample Accuracy was tracked using a sliding window approach. Subset Accuracy (also known as Exact Match) measures the proportion of instances where all predicted labels match exactly, whereas Sample Accuracy reflects partial correctness using the \textit{Jaccard} index per instance.  Let \( Y_i = \{y_{i1}, \dots, y_{iL}\} \) and \( Z_i = \{z_{i1}, \dots, z_{iL}\} \) be the true and predicted label sets for the \( i \)-th instance, respectively, with \(L\) labels and \(n\) instances, the metrics are:
\vspace{-5pt}

\begin{equation}
\text{Subset Accuracy} = \frac{1}{n} \sum_{i=1}^{n} \mathbf{1}[Y_i = Z_i]
\end{equation}
\vspace{-10pt}

\begin{equation}
\text{Hamming Loss} = \frac{1}{nL} \sum_{i=1}^{n} \sum_{l=1}^{L} \mathbf{1}[y_{il} \neq z_{il}]
\end{equation}
\vspace{-10pt}

\begin{equation}
\text{Macro F1} = \frac{1}{L} \sum_{l=1}^{L} \frac{2 \cdot \sum_{i=1}^{n} \mathbf{1}[y_{il} = 1 \wedge z_{il} = 1]}{\sum_{i=1}^{n} \mathbf{1}[y_{il} = 1] + \sum_{i=1}^{n} \mathbf{1}[z_{il} = 1]}
\end{equation}
\vspace{-10pt}

\begin{equation}
\text{Micro F1} = \frac{2 \cdot \sum_{l=1}^{L} \sum_{i=1}^{n} \mathbf{1}[y_{il} = 1 \wedge z_{il} = 1]}{\sum_{l=1}^{L} \sum_{i=1}^{n} \mathbf{1}[y_{il} = 1] + \sum_{l=1}^{L} \sum_{i=1}^{n} \mathbf{1}[z_{il} = 1]}
\end{equation}

\textbf{Datasets.} 
Table \ref{tab:real_datasets} provides a detailed overview of the real-world datasets used in this study, which are publicly available and widely used in multi-label classification tasks. These datasets range from small to large-scale, with up to 269,648 instances, 1836 features, and 374 labels, making them highly relevant for evaluating the scalability of iHOMER in high-dimensional streams. Additionally, the table includes key multi-label statistics such as \textit{cardinality} (the average number of labels per instance), \textit{density} (cardinality divided by the total number of labels), and the \textit{mean imbalance ratio} (the average degree of label imbalance). The last column of the table indicates whether the instances in each dataset are presented in a temporal order: yes ($\checkmark$ ), no ($\times$ ), or unknown (-). While these datasets do not explicitly provide information about concept drift, the temporal ordering can offer insight into the potential for concept drift.

\begin{table}[H]
\centering
\caption{Real-world datasets used. Temp. = Temporally ordered.}
\renewcommand{\arraystretch}{0.8}
\resizebox{0.48\textwidth}{!}{%
\begin{tabular}{lrrrrrrr}
\toprule
\textbf{Dataset} & \textbf{Inst.} & \textbf{Feat.} & \textbf{Lbls} & \textbf{Card.} & \textbf{Dens.} & \textbf{MeanIR} & \textbf{Temp.} \\
\midrule
    Flags & 194 & 19 & 7 & 3.39 & 0.48 & 2.255 & $\times$ \\
    WaterQuality & 1.060 & 16 & 14 & 5.07 & 0.36 & 1.767 & - \\
    Emotions & 593 & 72 & 6 & 1.87 & 0.31 & 1.478 & $\times$ \\
    VirusGO & 207 & 749 & 6 & 1.22 & 0.20 & 4.041 & $\times$ \\
    Birds & 645 & 260 & 19 & 1.01 & 0.05 & 5.407 & - \\
    Yeast & 2.417 & 103 & 14 & 4.24 & 0.30 & 7.197 & $\times$ \\
    Scene & 2.407 & 294 & 6 & 1.07 & 0.18 & 1.254 & $\times$ \\
    CAL500 & 502 & 68 & 174 & 26.04 & 0.15 & 20.578 & $\times$ \\
    Human & 3.106 & 440 & 14 & 1.19 & 0.08 & 15.289 & - \\
    Yelp & 10.806 & 671 & 5 & 1.64 & 0.33 & 2.876 & $\checkmark$ \\
    Medical & 978 & 1.449 & 45 & 1.25 & 0.03 & 89.501 & $\checkmark$ \\
    Eukaryote & 7.766 & 440 & 22 & 1.15 & 0.05 & 45.012 & - \\
    Slashdot & 3.782 & 1.079 & 22 & 1.18 & 0.05 & 19.462 & $\checkmark$ \\
    Hypercube & 100.000 & 100 & 10 & 1.00 & 0.10 & - & - \\
    Hypersphere & 100.000 & 100 & 10 & 2.31 & 0.23 & - & - \\
    Langlog & 1.460 & 1.004 & 75 & 15.94 & 0.21 & 39.267 & $\checkmark$ \\
    Stackex & 1.675 & 585 & 227 & 2.41 & 0.01 & 85.790 & $\checkmark$ \\
    Tmc & 28.596 & 500 & 22 & 2.22 & 0.10 & 17.134 & $\checkmark$ \\
    Ohsumed & 13.929 & 1.002 & 23 & 0.81 & 0.04 & 7.869 & $\checkmark$ \\
    D20ng & 19.300 & 1.006 & 20 & 1.42 & 0.07 & 1.007 & $\checkmark$ \\
    Bibtex & 7.395 & 1.836 & 159 & 2.40 & 0.02 & 12.498 & $\times$ \\
    Nuswidec & 269.648 & 129 & 81 & 1.87 & 0.02 & 95.119 & - \\
    Imdb & 120.919 & 1.001 & 28 & 1.00 & 0.04 & 25.124 & $\times$ \\
   
\bottomrule
\end{tabular}}
\label{tab:real_datasets}
\end{table}

\textbf{Average results.} Table \ref{tab:average_metrics_real_transposed} presents the average scores across four key evaluation metrics. Among all evaluated models, iHOMER achieves the best overall results on three of the four metrics—Subset Accuracy, Micro F1, and Hamming Loss—demonstrating its ability to adaptively model label dependencies through dynamic label clustering. This hybrid global-local architecture effectively captures co-occurrence patterns that static approaches miss. Compared to both global and local baselines, iHOMER shows consistent and substantial gains, particularly in Subset Accuracy (0.387) and Micro F1 (0.553). The closest competitor in Macro F1 is MLHAT, a global method, which outperforms iHOMER in that metric (0.445 vs. 0.282), likely due to a more balanced treatment of rare labels. In contrast, local models, which treat labels or label subsets independently, generally perform worse, especially on metrics sensitive to inter-label dependencies, such as Subset Accuracy and Micro F1. Nevertheless, local learners such as EFDT, HAT, kNN, and AMR emerged as the most competitive within their category. While these models demonstrate some capacity to model local patterns, they fall short when label relationships become complex or shift over time.

\begin{table}[htbp]
\renewcommand{\arraystretch}{0.7}
\centering
\caption{Avg. metric scores across datasets. Type = \textbf{G}lobal or \textbf{L}ocal.}
\resizebox{0.48\textwidth}{!}{%
% \begin{tabular}{lccccc}
% \begin{tabular}{lcccc@{\hskip 3pt}c}
\begin{tabular}{l@{\hskip -1pt}ccccc}
\toprule
\textbf{Method} & \textbf{Subset Acc.} & \textbf{Micro F1} & \textbf{Macro F1} & \textbf{H. Loss} & \textbf{Type} \\
\midrule
iHOMER   & \textbf{0.387} & \textbf{0.553} & 0.282 & \textbf{0.093} & G+L \\
MLHAT    & 0.351 & 0.550 & \textbf{0.445} & 0.105 & G \\
MLBELS   & 0.281 & 0.502 & 0.326 & 0.123 & G \\
MLHT     & 0.178 & 0.259 & 0.136 & 0.160 & G \\
MLHTPS   & 0.175 & 0.298 & 0.208 & 0.132 & G \\
iSOUPT   & 0.269 & 0.436 & 0.303 & 0.114 & G \\
KNN      & 0.275 & 0.430 & 0.303 & 0.111 & L \\
NB      & 0.133 & 0.301 & 0.225 & 0.156 & L \\
AMR      & 0.267 & 0.428 & 0.312 & 0.113 & L \\
HT      & 0.204 & 0.385 & 0.257 & 0.116 & L \\
HAT      & 0.240 & 0.423 & 0.287 & 0.110 & L \\
EFDT     & 0.230 & 0.434 & 0.299 & 0.110 & L \\
SGT      & 0.049 & 0.199 & 0.153 & 0.300 & L \\
MT       & 0.171 & 0.288 & 0.188 & 0.123 & L \\
ARF      & 0.260 & 0.417 & 0.288 & 0.098 & L \\
AMF      & 0.243 & 0.403 & 0.281 & 0.101 & L \\
ABALR    & 0.279 & 0.435 & 0.318 & 0.109 & L \\
ABOLR   & 0.277 & 0.433 & 0.319 & 0.110 & L \\

\bottomrule
\end{tabular}
}

\label{tab:average_metrics_real_transposed}
\end{table}

\begin{figure}[h]
    \centering
    \includegraphics[width=0.45\textwidth]{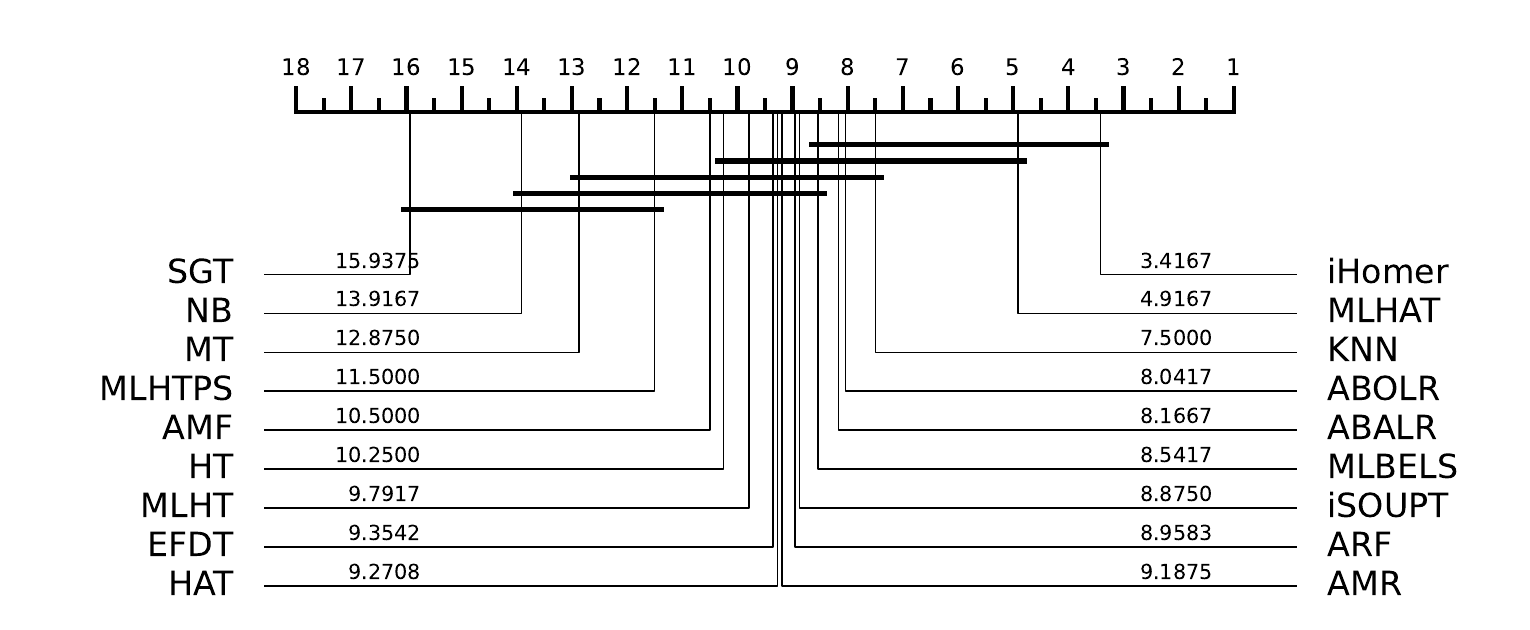}
    \caption{Nemenyi test - Subset Accuracy.}
    \label{fig:cdd_acc}
    \vspace{0.4cm}
\end{figure}

\begin{figure}[h]
    \centering
    \includegraphics[width=0.45\textwidth]{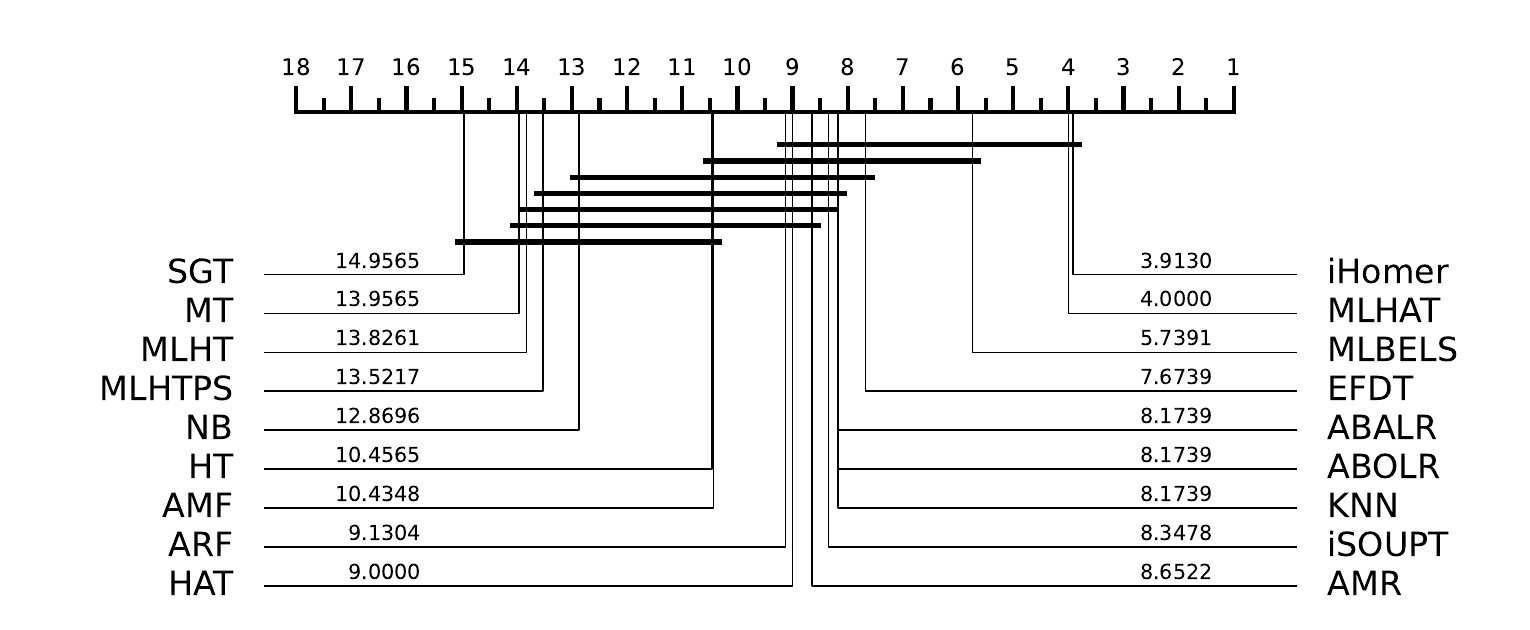}
    \caption{Nemenyi test - Micro F1.}
    \label{fig:cdd_microf1}
    \vspace{0.6cm}
\end{figure}

\begin{figure}[h]
    \centering
    \includegraphics[width=0.45\textwidth]{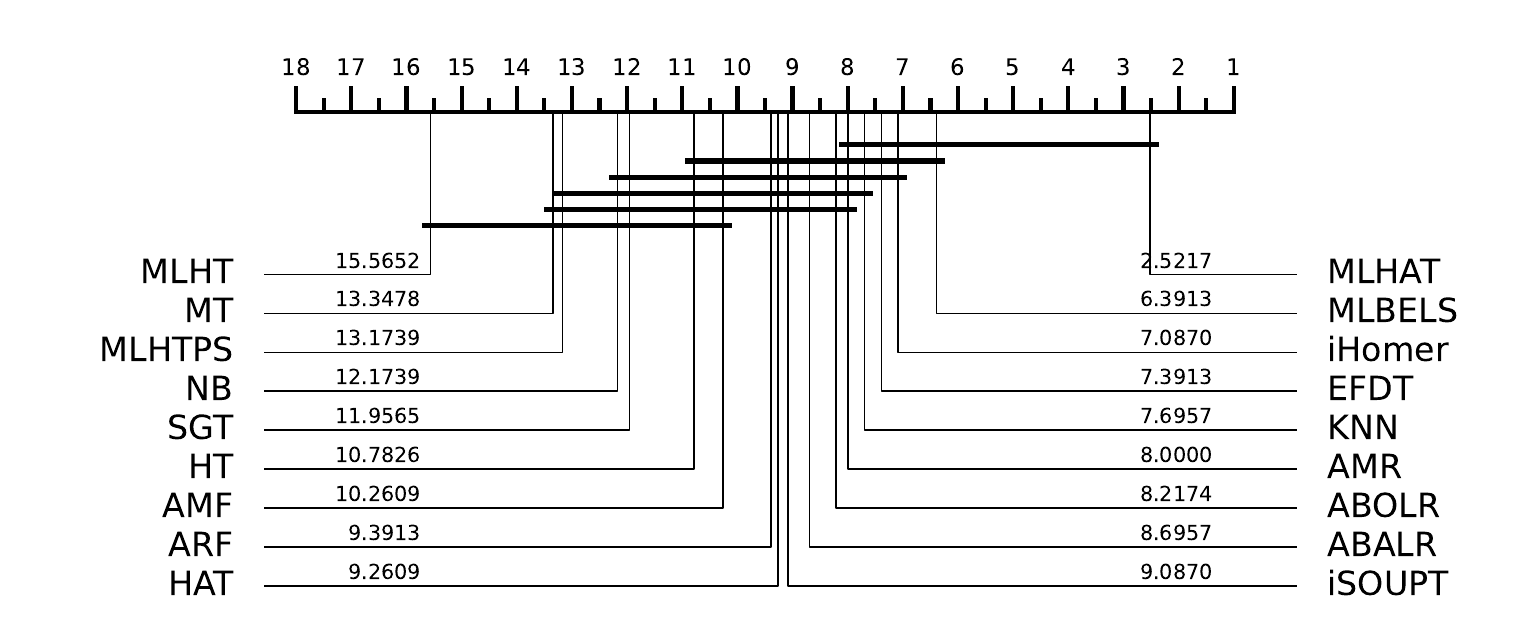}
    \caption{Nemenyi test - Macro F1.}
    \label{fig:cdd_macrof1}
    \vspace{0.25cm}
\end{figure}

\begin{figure}[h]
    \centering
    \includegraphics[width=0.45\textwidth]{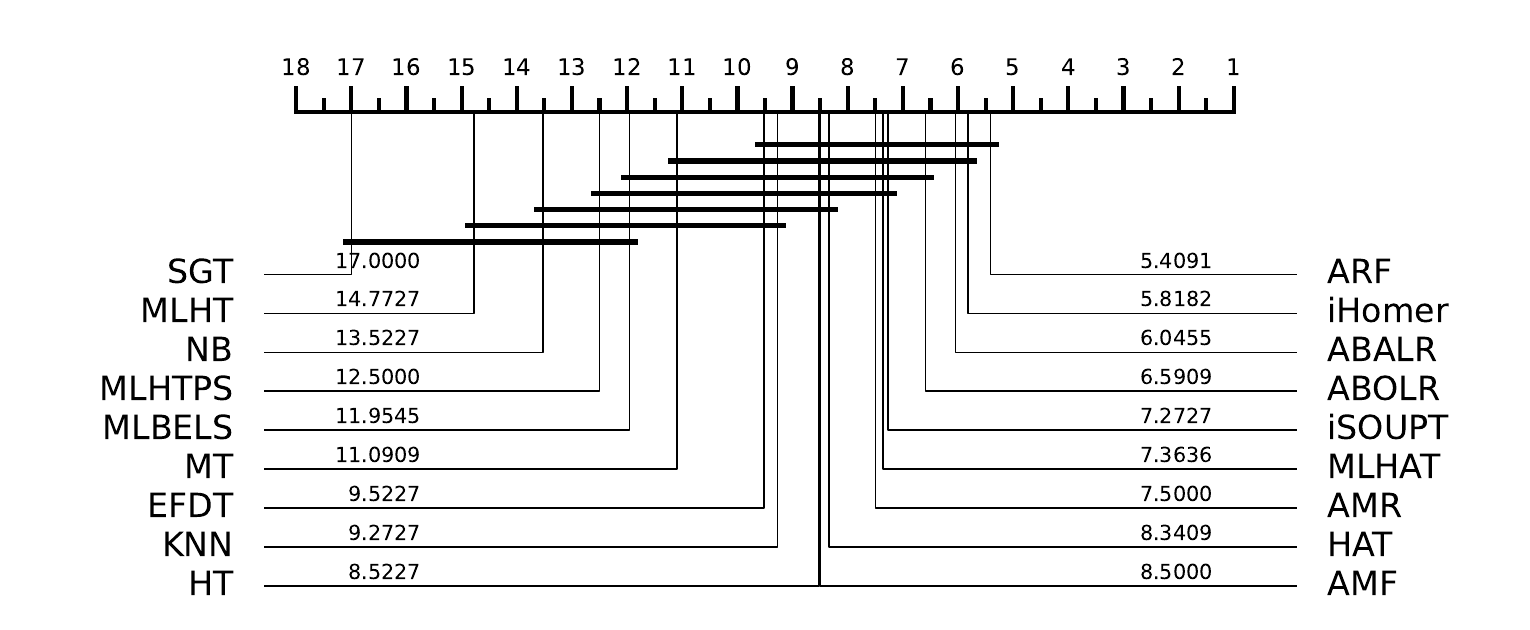}
    \caption{Nemenyi test - Hamming Loss.}
    \label{fig:cdd_hamming}
    \vspace{0.6cm}
\end{figure}

\textbf{Critical difference diagrams.} To validate the statistical significance of metric scores differences, the Friedman test was conducted, with the resulting statistics ranging between approximately 122 and 148 for all metrics. In all cases, the p-value was zero, indicating a clear rejection of the null hypothesis that all methods perform equally. A Nemenyi post-hoc test was then applied, with the resulting Critical Difference Diagrams (CDDs) for 95\% confidence in Figures \ref{fig:cdd_acc}, \ref{fig:cdd_microf1}, \ref{fig:cdd_macrof1}, and \ref{fig:cdd_hamming}, further corroborating its consistent performance across all datasets. Notably, iHOMER ranks among the top three methods in all CDDs, underscoring its robustness across both label-based and instance-based performance criteria. In Subset Accuracy and Micro F1, iHOMER holds a statistically significant lead over most local learners and several global ones, affirming its effectiveness in capturing high-fidelity label combinations and optimizing overall predictive balance. While iHOMER remains highly competitive in Hamming Loss, it does not claim the top position. This outcome is partly attributable to the metric’s design: Hamming Loss penalizes every individual label misclassification equally, regardless of whether the rest of the label set is correctly predicted. As a hybrid model that prioritizes structured label dependencies and exact match accuracy, iHOMER optimizes for holistic correctness rather than minimizing marginal errors. Consequently, it achieves higher Subset Accuracy and F1 scores at the expense of slightly increased label-wise mismatches, which are more harshly reflected in Hamming Loss.

\begin{figure*}[h!]
    \centering
    \includegraphics[width=0.95\linewidth]{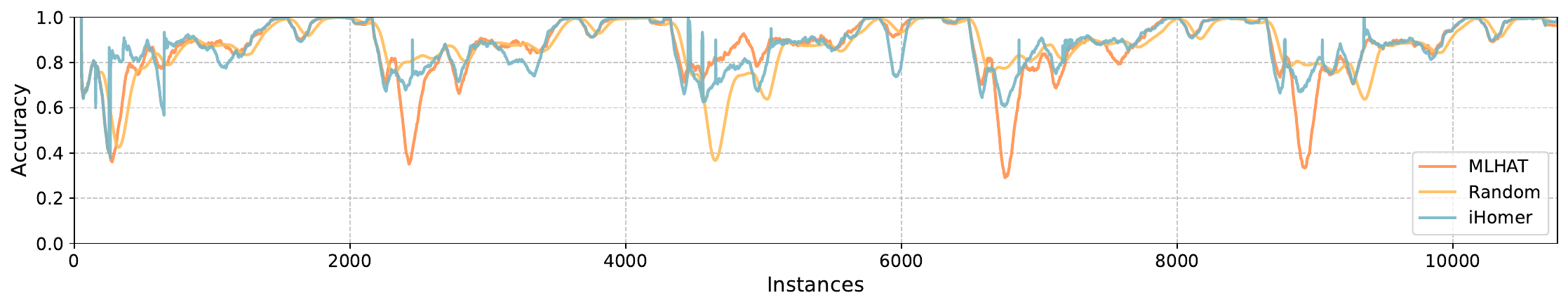}
    \caption{Rolling Sample Accuracy on the Yelp dataset.}
    \label{fig:rolling}
\end{figure*}

\textbf{iHOMER vs. MLHAT.} 
While MLHAT is recognized in the literature as a strong multi-label classifier, iHOMER consistently outperforms it across diverse scenarios involving varying label cardinality, density, and concept drift. These results highlight iHOMER’s ability to accurately model complex label dependencies and capture complete label combinations. As shown in Table \ref{tab:average_metrics_real_transposed}, iHOMER achieves a substantial 10.26\% improvement in Subset Accuracy (widely regarded as the most rigorous evaluation metric for multi-label tasks) and an 11.43\% reduction in Hamming Loss compared to MLHAT. These gains indicate both higher predictive accuracy and fewer misclassifications. iHOMER also shows a modest improvement in Micro F1, while exhibiting slightly lower performance on Macro F1. This drop is expected, as iHOMER clusters disjoint label subsets, which may underrepresent rare labels when they are isolated into separate groups. Both models benefit from using multi-label metrics for drift detection, enabling quicker adaptation than methods that handle each label independently. Wilcoxon signed-rank tests confirm that the improvements in Subset Accuracy and Hamming Loss are statistically significant. Although iHOMER also achieves higher average Micro F1 scores, this particular difference does not reach statistical significance. To illustrate iHOMER’s superiority on a representative dataset, Figure \ref{fig:rolling} presents the rolling Sample Accuracy over time, highlighting the comparative performance of iHOMER, a single MLHAT, and MLHAT ensembles with randomly generated hybrid partitions. The rolling metric reveals that iHOMER produces smoother, more stable trends and avoids the sharp drops observed in other methods, underscoring the value of informed label space partitioning in streaming data scenarios.

\begin{table}[h!]
\centering
\caption{Wilcoxon signed-rank test: iHOMER vs. MLHAT.}
\begin{tabular}{l c c c}
\toprule
\textbf{Metric}        & \textbf{Statistic} & \textbf{P-value} & \textbf{P-value < 0.05} \\ 
\midrule
Subset Acc.   & 72.0000            & 0.0249                    &     \checkmark                    \\
Micro F1      & 114.0000           & 0.4820                    &                       \\
Macro F1      & 21.0000            & 0.0001                    &   \checkmark           \\
Hamming Loss  & 20.0000            & 0.0002                    &   \checkmark          \\
\bottomrule
\end{tabular}%
\label{tab:wilcoxon_results_real}
\end{table}

\section{Conclusions}

In this paper, iHOMER (Incremental Hierarchy Of Multi-label Classifiers) was introduced for streaming multi-label learning. iHOMER incrementally partitions the label space into disjoint, correlated clusters without relying on predefined hierarchies. iHOMER leverages online divisive-agglomerative clustering based on \textit{Jaccard} similarity and a global tree-based learner driven by a multivariate \textit{Bernoulli} process to guide instance partitioning. To address non-stationarity, it integrates drift detection mechanisms at both global and local levels, enabling dynamic restructuring of label partitions and subtrees. Experiments across 23 benchmarks show that iHOMER outperforms state-of-the-art baselines by more than 20\% and outperforms random search by 40\%. These findings suggest that conventional multi-label streaming methods may not fully capture label correlations, and clustering the label space offers a promising solution to addressing challenges such as high dimensionality, class imbalance, and the complexity of label correlations.

\textbf{Future work.} Despite its strengths, iHOMER still presents opportunities for improvement in computational efficiency. While it inherits MLHAT's ability to respond to concept drift via node resplitting and information gain approximation \cite{esteban2024hoeffding}, as well as ODAC's dynamic dendrogram expansion based on cluster diameters \cite{odac}, its growth efficiency remains suboptimal. Potential enhancements include dynamically tuning parameters such as \textit{ grace periods}, \textit{tie thresholds}, and \textit{split criteria} in response to incoming data. Further improvements could involve implementing stricter split evaluation rules (based on entropy, information gain, and minimum leaf instance counts), adaptive expansion strategies, and memory-aware mechanisms like leaf deactivation. These changes would allow more computational focus on frequently accessed nodes while reducing overhead elsewhere \cite{lourencco2025device}. Finally, testing iHOMER on a broader range of datasets with annotated drift events could provide deeper insights into its adaptability and robustness.

%%%%%%%%%%%%%%%%%%%%%%%%%%%%%%%%%%%%%%%%%%%%%%%%%%%%%%%%%%%%%%%%%%%%%%%%

%%% Use this environment to include acknowledgements (optional).
%%% This will be omitted in doubleblind mode.

\begin{ack}
Work funded by Portuguese Foundation for Science and Technology under project doi.org/10.54499/UIDP/00760/2020, Ph.D. scholarship doi.org/10.54499/PRT/BD/154713/2023, and Luso-American Development Foundation (FLAD Proj. 2025-0018). It also received EU funds, through Portuguese Republic’s Recovery and Resilience Plan, within project PRODUTECH R3.
\end{ack}

%%%%%%%%%%%%%%%%%%%%%%%%%%%%%%%%%%%%%%%%%%%%%%%%%%%%%%%%%%%%%%%%%%%%%%%%

%%% Use this command to include your bibliography file.

\bibliography{mybibfile}

\end{document}